\documentclass[10pt,twocolumn,letterpaper]{article}

\usepackage{iccv}
\usepackage{times}
\usepackage{epsfig}
\usepackage{graphicx}
\usepackage{amsmath}
\usepackage{amssymb}

\usepackage{color}
\usepackage{subfigure}
\usepackage{dsfont}
\usepackage{tabularx}
\usepackage{booktabs}
\usepackage[small]{caption}

\usepackage[pagebackref=true,breaklinks=true,letterpaper=true,colorlinks,bookmarks=false]{hyperref}

\newcommand{\comment}[1]{}

\newcommand{\bmu}{\boldsymbol{\mu}}

\newcommand{\bI}{\mathbf{I}}
\newcommand{\tbI}{\tilde{\mathbf{I}}}

\newcommand{\bsigma}{\boldsymbol{\sigma}}

\newcommand{\bX}{\mathbf{X}}
\newcommand{\bY}{\mathbf{Y}}

\newcommand{\bx}[0]{\mathbf{x}}
\newcommand{\tbx}[0]{\tilde{\mathbf{x}}}

\newcommand{\by}[0]{\mathbf{y}}

\newcommand{\calN}{\mathcal{N}}

\newcommand{\IR}{\mathds{R}}

\newcommand{\bA}{\mathbf{A}}

\newcommand{\fig}[1]{Fig.~\ref{fig:#1}}

\newcommand{\tbl}[1]{Table~\ref{tbl:#1}}
\newcommand{\secref}[1]{Section~\ref{sec:#1}}
\newcommand{\refsec}[1]{Section~\ref{sec:#1}}

\newcommand{\transform}{T}

\definecolor{orange}{rgb}{1,0.5,0}

\newcommand{\eq}[1]{Eq.~\eqref{eq:#1}}

\usepackage{soul}

\newcommand{\tu}{\tilde{u}}
\newcommand{\tv}{\tilde{v}}

\newcommand{\MNIST}{MNIST\xspace}
\newcommand{\PASCAL}{Pascal VOC\xspace}
\newcommand{\TRAFFIC}{GTSRB\xspace}

\definecolor{darkolivegreen}{rgb}{0.3, 0.6, 0.1}

\newcommand{\bE}{\mathbf{E}}

\newcommand{\btheta}{\boldsymbol{\theta}}

\iccvfinalcopy 

\def\iccvPaperID{811} 

\ificcvfinal\fi
\begin{document}

\title{Linearized Multi-Sampling for Differentiable Image Transformation}

\author{Wei Jiang\textsuperscript{1} \qquad Weiwei Sun\textsuperscript{1} \qquad Andrea Tagliasacchi\textsuperscript{1,2} \qquad Eduard Trulls\textsuperscript{2} \qquad Kwang Moo Yi\textsuperscript{1}
\vspace{0.5em}
\\
{\small \textsuperscript{1}Visual Computing Group, University of Victoria\qquad
\textsuperscript{2}Google Research}\\
{\tt\small \{jiangwei, weiweisun, kyi\}@uvic.ca, \{taglia, trulls\}@google.com}
}

\twocolumn[{%
\renewcommand\twocolumn[1][]{#1}%
\maketitle
\begin{center}
\vspace{-0.7cm}
\centering
\includegraphics[width=\linewidth]{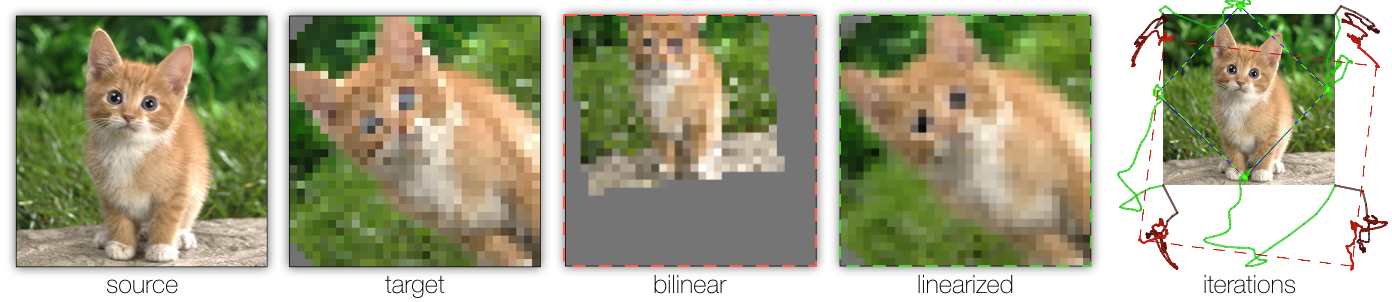}
\vspace{-0.4cm}
\captionof{figure}{
{\bf Image alignment example -}
We iteratively 
find the image transformation parameters that produce the downsampled \textit{target} image when applied to the \textit{source} image using different sampling strategies.
We visualize 
the transformed source image
at convergence when using {\bf {\color{red}{\textit{bilinear}}}} sampling and the proposed {\bf {\color{darkolivegreen}{\textit{linearized}}}} sampling.
We further display the optimization path -- \textit{iterations} -- for each method.
The proposed linearized sampling provides improved gradients, leading to better convergence.
}
\label{fig:teaser}
\end{center}%
}]

\begin{abstract}
We propose a novel image sampling method for differentiable image transformation in deep neural networks.
The sampling schemes currently used in deep learning, such as Spatial Transformer Networks, rely on bilinear interpolation, which performs poorly under severe scale changes, and more importantly, results in poor gradient propagation. 
This is due to their strict reliance on direct neighbors. 
Instead, we propose to generate random auxiliary samples in the vicinity of each pixel in the sampled image, and create a linear approximation with their intensity values. 
We then use this approximation as a differentiable formula for the transformed image. 

We demonstrate that our approach produces more representative gradients with a wider basin of convergence for image alignment, which leads to considerable performance improvements when training networks for classification tasks.
This is not only true under large downsampling, but also when there are no scale changes. 
We compare our approach with multi-scale sampling and show that we outperform it.
We then demonstrate that our improvements to the sampler are compatible with other tangential improvements to Spatial Transformer Networks and that it further improves their performance.
\footnote{
Code and models are available at \url{https://github.com/vcg-uvic/linearized_multisampling_release}.
}
\end{abstract}

\section{Introduction}
The seminal work of~\cite{Jaderberg15} introduced Spatial Transformer Networks (STN), a differentiable component that allows for spatial manipulation of image data by deep networks.
It has since become commonplace to include attention mechanisms in deep architectures in the form of image transformation operations.
STNs have been applied to object detection \cite{Dai17a}, segmentation~\cite{He17, Liu2018}, dense image captioning~\cite{Johnson16a}, local correspondence for image classification~\cite{Altwaijry16}, local features~\cite{Yi16a, Ono18,Mishkin18}, and as a tool for local hard attention~\cite{Kuen16}.
Regardless of application and architecture, all of these methods rely on bilinear interpolation.

A major drawback of bilinear interpolation is that it is \textit{extremely} local -- it considers only the four closest pixel neighbors of the query.
As this sampling does not account for the magnitude of the applied transformation, the performance of networks that rely on it
degrades when scale changes are severe, as for example shown in \fig{teaser}.
This shortcoming of the differentiable sampler was already hinted at in the original STN paper~\cite{Jaderberg15}, but never fully investigated.
Note that this is a problem in practice, as applications that leverage attention mechanisms often transform the image to a resolution much lower than the original~\cite{Yi16b,Ono18,He17}.
Furthermore, STNs are typically used as pre-alignment networks for classification~\cite{Jaderberg15,Lin16b}.
The inability of the bilinear sampler in the STN to cope with large downsampling results in increased capacity requirements for the classification network -- this extra capacity will be needed to learn invariance to transformations the STN was not able to capture.
We show that this is indeed the case by demonstrating that a better classification accuracy can be achieved with the \emph{same} network by replacing the sampling operation with ours.

Several methods have been proposed to improve the stability and robustness of networks that employ bilinear sampling.
For example, Lin \etal.~\cite{Lin16b} introduced inverse compositional spatial transformer networks (ICSTN), which decompose transformations into smaller ones.
Jia \etal.~\cite{Jia17} proposed to warp features rather than images. Shu \etal.~\cite{Shu18} proposed a hierarchical strategy incorporating flow-fields.
However, all of these methods still rely on heavily localized bilinear interpolation when it comes to computing the values of individual pixels.

Instead, we take our inspiration from the well-known Lucas-Kanade (LK) optical flow algorithm~\cite{Lucas81,Baker04b}, and linearize the interpolation process by building a suitable first-order approximation by randomly sampling the neighborhood of the query.
These auxiliary samples are created in the \textit{transformed} image domain, so that their spatial locations take into account the transformation applied to the image.
In other words, the resulting gradients are \textit{transformation-aware}, and able to capture how the image changes according to the transformation.

However, this process can cause sample locations to collapse, as the transformation may warp them all to a single pixel -- for example, when the transformer zooms into to a small area of the image.
To overcome this issue, we propose an effective solution that perturbs the auxiliary samples after warping.
This two-stage strategy allows our approach to deal with \textit{any} type of affine transformation.

Our experiments demonstrate that this allows for a wider basin of convergence for image alignment, which helps the underlying optimization task.
Most importantly, any network with embedded attention mechanisms, \ie anywhere differentiable image transformation is required, should perform better when substituting standard bilinear interpolation with our approach.
We demonstrate this by adding our sampling method to ICSTN~\cite{Lin16b}, and show that we are able to not only reduce its error rate by 14.6\% (relative) even \emph{without} down sampling, but also that we effectively eliminate the harming effect of 4x downsampling on the performance of the classifier (error at 1x is $4.85\%$ and at 4x is $4.86\%$).
This means that we can use a classification network that is 11 times smaller than the original, without any performance loss.

\section{Related work}
\def \relworkvspace {-1.0em}

There is a vast amount of work in the literature on estimating spatial transformations, with key applications to image registration problems, going back to the pioneering work of Lucas and Kanade~\cite{Lucas81}. 
Here we review previous efforts on image alignment and sampling techniques, particularly with regards to their use in deep learning.

\vspace{\relworkvspace}
\paragraph{Linearization.}
The Lucas \& Kanade (LK) algorithm~\cite{Lucas81} predicts the transformation parameters using linear regression.
It mathematically linearizes the relationship between pixel intensities and pixel locations by taking the first-order Taylor approximation. 
In this manner, the sampling process can be enhanced by enforcing linear relationships during the individual pixel sampling.
This approach has proved greatly successful in many applications such as optical flow estimation~\cite{Barron94}.
Linearization is also widely utilized to improve the consistency of the pixel values in image filtering~\cite{he13}.

\vspace{\relworkvspace}
\paragraph{Multisampling.}
Multisampling is a common approach to enhance the reliability of sampling strategies.
For example, one can sample multiple nearby pixels before feeding them to a classifier, jointly smoothing the scores from 
a pixel and its neighbors~\cite{Zhang14d,Chen11c}.
Non-local means~\cite{Buades05} computation can be sped up by Markov Chain Monte Carlo (MCMC) sampling~\cite{Chen14}.
Sampling can also be employed to compute finite differences at large stencils to produce gradient estimates that are less sensitive to noise and discontinuities~\cite{Tan16}.

\vspace{\relworkvspace}
\paragraph{In the context of deep learning.}
Early efforts in deep learning for computer vision were limited by their inability to manipulate input data, a crucial requirement in achieving spatial invariance.
Jaderberg \etal. proposed to address this shortcoming with Spatial Transformer Networks~\cite{Jaderberg15}, which introduced the concept of a differentiable transformation to actively manipulate the input image or the feature maps.
This effectively enabled learning hard attention mechanisms in and end-to-end fashion. 
To achieve this, they introduced a differentiable sampling operation, making it possible to propagate the loss in subsequent tasks with respect to the parameters of the predicted transformation.

STNs are widely used in applications that operate on image patches~\cite{Wu17a, Shu18, Zhang17e, bhagavatula17e}.
Modern methods, such as Wang \etal.~\cite{Wang18e}, improve the patch sampling through an in-network transformation.
Qi \etal. introduced PointNet~\cite{Qi17a}, a deep network for 3D point cloud segmentation which relies on STN to learn to transform the data into a canonical form before giving it to the network.
Nevertheless, the paper reports only marginal performance improvements with the use of the STN, which demonstrates there is potential for further research in this area.

Several methods use bilinear sampling without explicitly learning a transformation.
LIFT~\cite{Yi16b} and LF-Net~\cite{Ono18} relied on STN to warp image patches and learn local features in an end-to-end manner, with the transformation parameters given by specially-tailored networks (\eg keypoint detection).
AffNet~\cite{Mishkin18} applied a similar strategy to learn affine-covariant regions. Polar Transformer Networks~\cite{Esteves18} were used by~\cite{Ebel19} to build scale-invariant descriptors by transforming the input patch into log-polar space.

Bilinear sampling has also been used in the context of image upsampling, depth estimation, and segmentation~\cite{Godard17,Chen18b,He17,Long15a,Ronneberger15}.
For example, in Mask R-CNN~\cite{He17}, the region of interest alignment layer mainly relies on bilinear sampling/interpolation.
These methods could also potentially benefit from enhanced sampling.

\vspace{\relworkvspace}
\paragraph{Current limitations of Spatial Transformers.}

Despite its popularity, in this paper we show how the bilinear sampler utilized in the STN framework is inherently unreliable and lacks robustness.
Several variants have been recently proposed in order to cope with these shortcomings.

Jia \etal~\cite{Jia17} 
showed how to exploit patterns found in the feature map deformations throughout the network layers that are insensitive to bilinear interpolation, thus boosting the accuracy of STN.
Chang \etal~\cite{chang17e} trained a deep network with a Lucas-Kanade layer to perform coarse-to-fine image alignment.
The inverse compositional network proposed in~\cite{Lin16b} passes the transformation parameters instead of the transformed image in the forward pass in order to mitigate errors in the bilinear sampling.
Recently, a hierarchical strategy was proposed to deal with transformations at multiple levels~\cite{Shu18}, by incorporating a U-Net~\cite{Ronneberger15}, providing richer transformation cues.
However, all of these still rely on bilinear sampling to transform the image.

\begin{figure}
\includegraphics[width=\linewidth, height=0.55\linewidth]{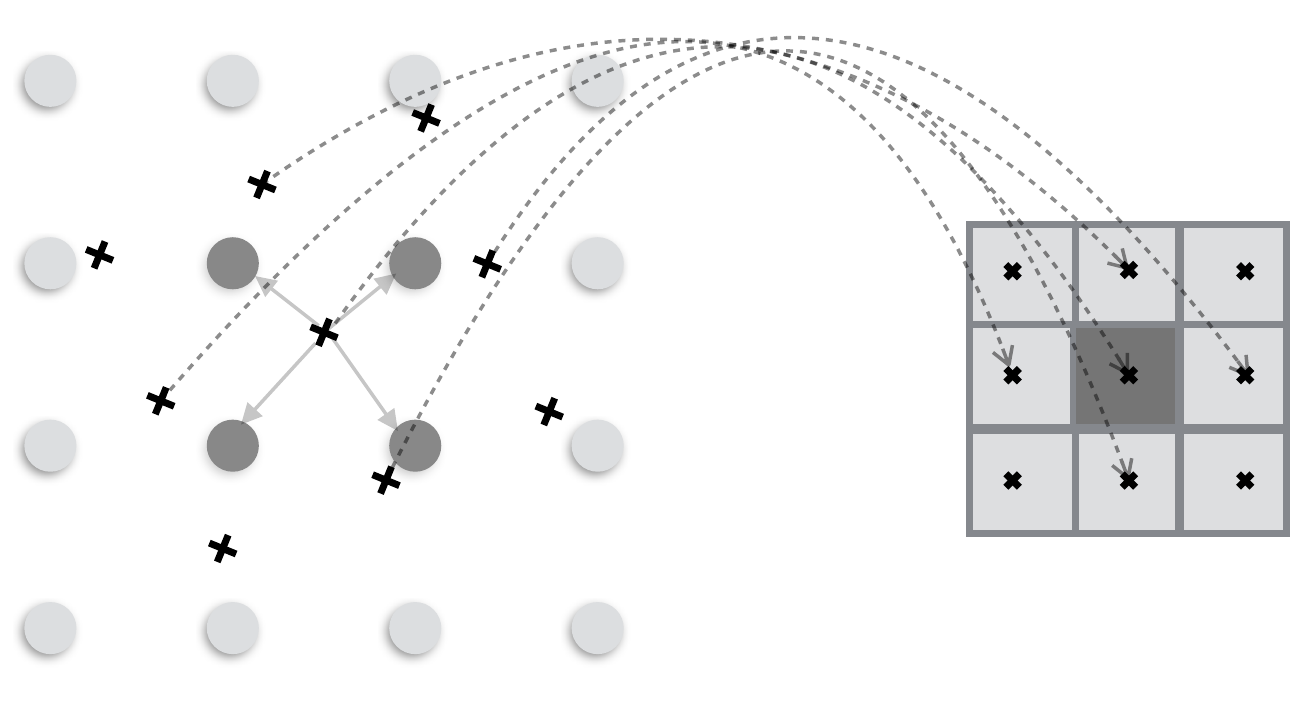}
\caption{
\textbf{Bilinear sampling -- }
To sample a $3\times3$ image on the right from the image on the left, each pixel is queried in order to find the exact corresponding point in the left image, according to a given transformation. 
As illustrated by the central pixel, the intensities 
are computed through the nearest neighbors by bilinear interpolation.
Note that, even when the pixels on the right fall into further away regions, the interpolation will {\em always} be performed from their direct neighbors.
}
\label{fig:prelim}
\end{figure}
\section{Differentiating through interpolation}
We now briefly review bilinear interpolation~\cite{Jaderberg15}; see \fig{prelim}.
The main reason that a form of interpolation -- typically bilinear interpolation -- is used to implement differentiable image transformations is that transformations require indexing operations, which are not differentiable by
themselves.
More specifically, 
denoting the image coordinates as $\bx\in\IR^2$, the image intensity at this coordinate as $\bI\left(\bx\right)\in\IR^{C}$, where $C$ is the number of channels in the image, and the coordinate transformation with parameters $\btheta$ as $\transform_{\btheta}\left(\bx\right)$,
then, the transformed image $\tilde{\bI}$ evaluated at $\tilde{\bx}$ is
\begin{equation}
    \tilde{\bI}\left(\tilde{\bx}\right) = \sum_{\bx}\bI\left(\transform_{\btheta}\left(\bx\right)\right) \: K\left(\tilde{\bx},\transform_{\btheta}\left(\bx\right)\right),
    \label{eq:interp}
\end{equation}
where $K\left(\cdot, \cdot\right)$ is the kernel which defines the influence of each pixel.
Note here that the image indexing operation $\bI\left(\transform_{\btheta}\left(\bx\right)\right)$ is non-differentiable, as it is a selection operation, and the way gradients propagate through the network depends on the kernel.
In theory, this kernel could leverage the entire image, thus making the gradient for all pixel values affect the optimization. However, this would require back-propagating through every pixel in the original image for every pixel in the transformed image, which is prohibitively expensive.
In the case of bilinear interpolation, the kernel is set so that $K(\bx, \by) = 0$ when $\bx$ and $\by$ are not direct neighbors.
Therefore, the gradient only flows through what we will refer to as \emph{sub-pixel gradients}, \ie, the intensity difference between neighboring pixels in the \emph{original} image.

This can be quite harmful under significant downsampling.
The sub-pixel gradients will not correspond to the large-scale changes that occur when the transformation parameter changes, as these cannot be captured by the immediate neighbourhood of a point.
In other words, the intensity values with non-zero gradients from kernel $K$ need to be chosen in a way that is invariant to these deformations, and carefully so that it reflects how the image would actually change under deformations.
We will now introduce a novel strategy to overcome this problem.

\begin{figure*}
\includegraphics[width=\linewidth]{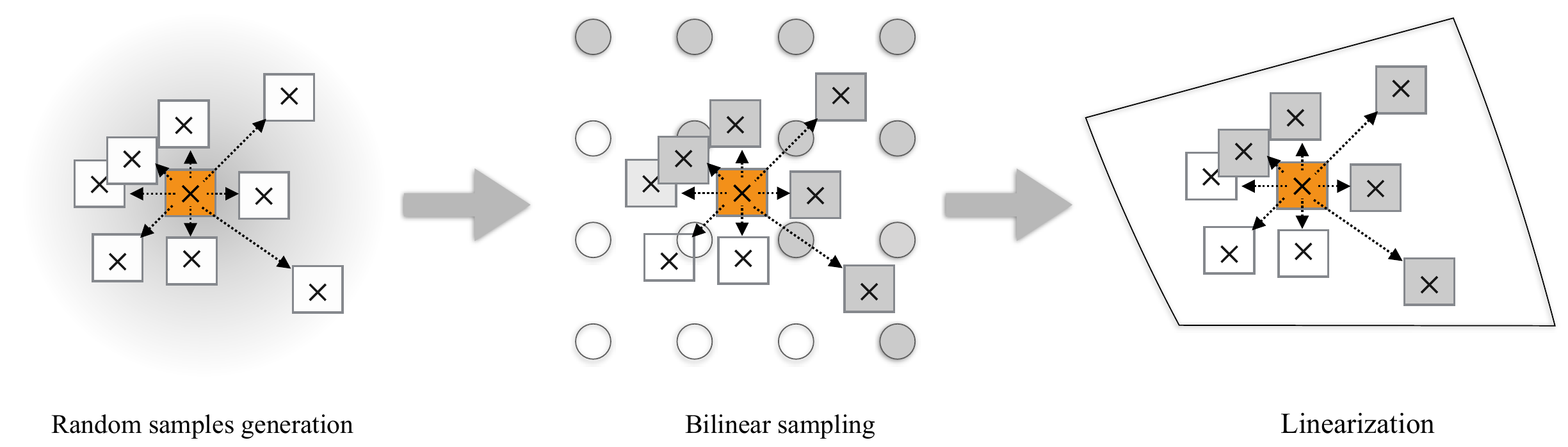}
\caption{
\textbf{Linearized multi-sampling -- }
For each pixel that we query, we generate a set of random auxiliary samples, whose intensities we extract through bilinear sampling.
We then process these intensities to create a linear approximation,
which we use as the differentiable representation for the intensity of the queried pixel.
}
\label{fig:framework}
\end{figure*}

\section{Linearized multi-sampling}

\fig{framework} illustrates our method, step by step.
Given a pixel location, we apply Gaussian noise to generate $K$ nearby auxiliary sample locations.
We then perform bilinear sampling at these locations.
After sampling, we use the intensities of these points and their coordinates to perform a linear approximation at the sample location.
Finally, these linear approximations are used as the differentiable representation of each pixel in the transformed image. This provides a larger context to the local transformation, and increases robustness under downsampling.

Formally, given the parameterization of the transformation $\btheta$, a desired sample point as $\bx_i$, where $i$ is the index for this sample, we write the linear approximation $\hat{\bI}\left(\bx\right)$ as
\begin{equation}
    \hat{\bI}\left(\bx\right) = 
    \bI\left(\transform_{\btheta}\left(\bx_i\right)\right) + 
    \bA_i \left(
        \transform_{\btheta}\left(\bx\right) - 
        \transform_{\btheta}\left(\bx_i\right)
    \right)
    ,
    \label{eq:linearization}
\end{equation}
where $\bA_i$ is a matrix that defines the linearization that we seek to find.
For the sake of explanation, let us assume for the moment that we have $\bA_i$.
Then, notice here that we are linearizing at the transformed coordinate $\transform_{\btheta}\left(\bx_i\right)$, thus treating everything except for $\transform_{\btheta}\left(\bx\right)$ as a constant.
Therefore, $\bA_i$ in \eq{linearization} corresponds to the gradient of $\hat{\bI}\left(\bx\right)$ with respect to $\bx$.

To obtain $\bA_i$ we first sample multiple points near
the desired sample points $\bx_i$ and find a least-square fit for the sample results.
Specifically,  we take $K$ samples
\begin{equation}
    \bx_i^{k} \sim \calN\left(\bx_i, \bsigma\right)
    ,\;\;
    \forall
    k \in \left\{1,2,...,K\right\}
    ,
    \label{eq:sample}
\end{equation}
where $\calN\left(\bmu,\bsigma\right)$ denotes a Gaussian distribution centered at $\bmu$ with standard deviation $\bsigma$, and $\bx_i^{0} = \bx_i$. 
In our experiments we set $\bsigma$ to match the pixel width and height of the sample output.
Note that by using Gaussian noise, we are effectively assuming a Gaussian point-spread function for each pixel when sampling.

We then obtain $\bA_i$ by least-squares fitting.
If we simplify the notation for $\tilde{\bI}\left(\transform_{\btheta}\left(\bx_i^{k}\right)\right)$ as $\tbI_i^{k}$, and for $\transform_{\btheta}\left(\bx_i^{k}\right)$ as $\tbx_i^{k}$, where $\tbx_i^k = \left[\tu_i^k, \tv_i^k\right]$, we form two data matrices $\bY_i$ and $\bX_i$, where 
\begin{equation}
   \bY_i = \bX_i\bA_i
   ,
\end{equation}
\begin{equation}
    \bY_i = 
    \begin{bmatrix}
        \tbI_i^{1} - \tbI_i^{0} &
        \tbI_i^{2} - \tbI_i^{0} &
        \cdots &
        \tbI_i^{K-1} - \tbI_i^{0}
    \end{bmatrix}^\top
    ,
    \label{eq:data_I}
\end{equation}
\begin{equation}
    \bX_i = 
    \begin{bmatrix}
        \tu_i^{1} - \tu_i^{0} & 
        \tu_i^{2} - \tu_i^{0} &
        &
        \tu_i^{K-1} - \tu_i^{0} \\
        \tv_i^{1} - \tv_i^{0} &
        \tv_i^{2} - \tv_i^{0} &
        \cdots &
        \tv_i^{K-1} - \tv_i^{0} \\
        1 & 1 &  & 1
    \end{bmatrix}^\top
    ,
    \label{eq:data_X}
\end{equation}
and then solve for A in a least square sense with Tikhonov regularization for numerical stability
\begin{equation}
    \bA_i = \left(\bX_i^\top\bX_i + \epsilon\bE\right)^{-1}\bX_i^\top\bY_i
    ,
    \label{eq:linearized_stable}
\end{equation}
where $\bE$ is the $3 \times 3$ identity matrix, and $\epsilon$ a small scalar.

\paragraph{Multi-scale sampling}
An alternative approach would be to use multi-scale alignment with auxiliary samples distributed over pre-defined grids at varying levels of coarseness.
This can scale up as much as desired -- potentially up to using the entire image as a neighborhood for each pixel -- but the increase in computational cost is linear with respect to the number of auxiliary samples.
Random selection allows us to capture both local and contextual
structure in an efficient manner.

\begin{figure}
\includegraphics[width=\linewidth, height=0.28\linewidth]{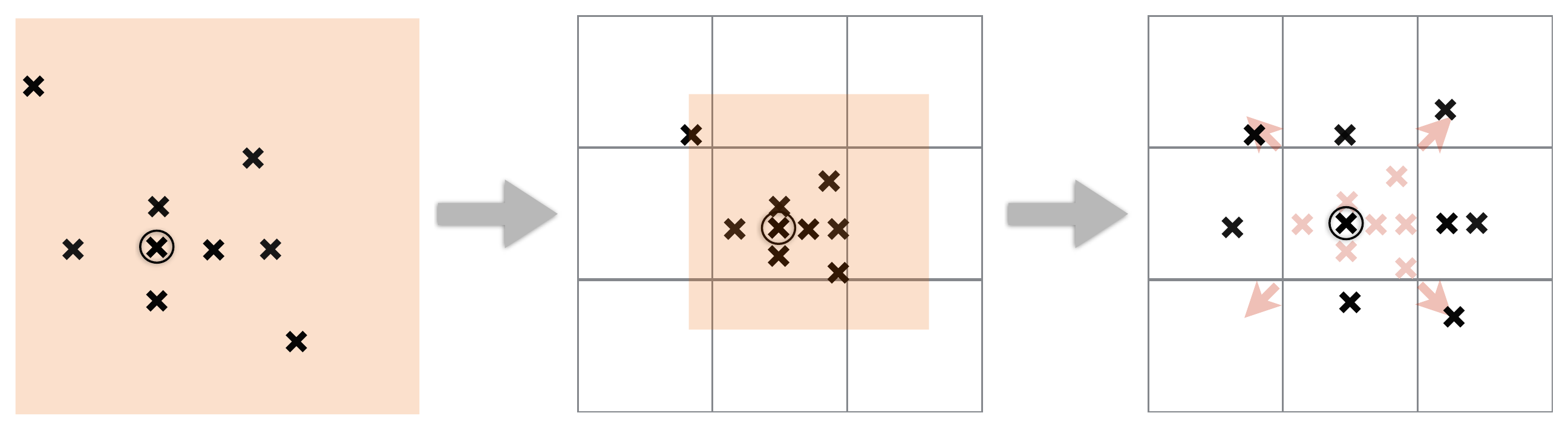}
\caption{
\textbf{Sample collapse prevention --}
During an upsampling operation, the auxiliary samples (left) could fall onto a single pixel of the sampled image once the transformation is applied (middle).
To prevent sample collapse, we apply additional noise to the {\em transformed} auxiliary sample locations (right).
}
\label{fig:pushaway}
\end{figure}
\begin{figure*}
\includegraphics[width=\linewidth, height=0.32105\linewidth]{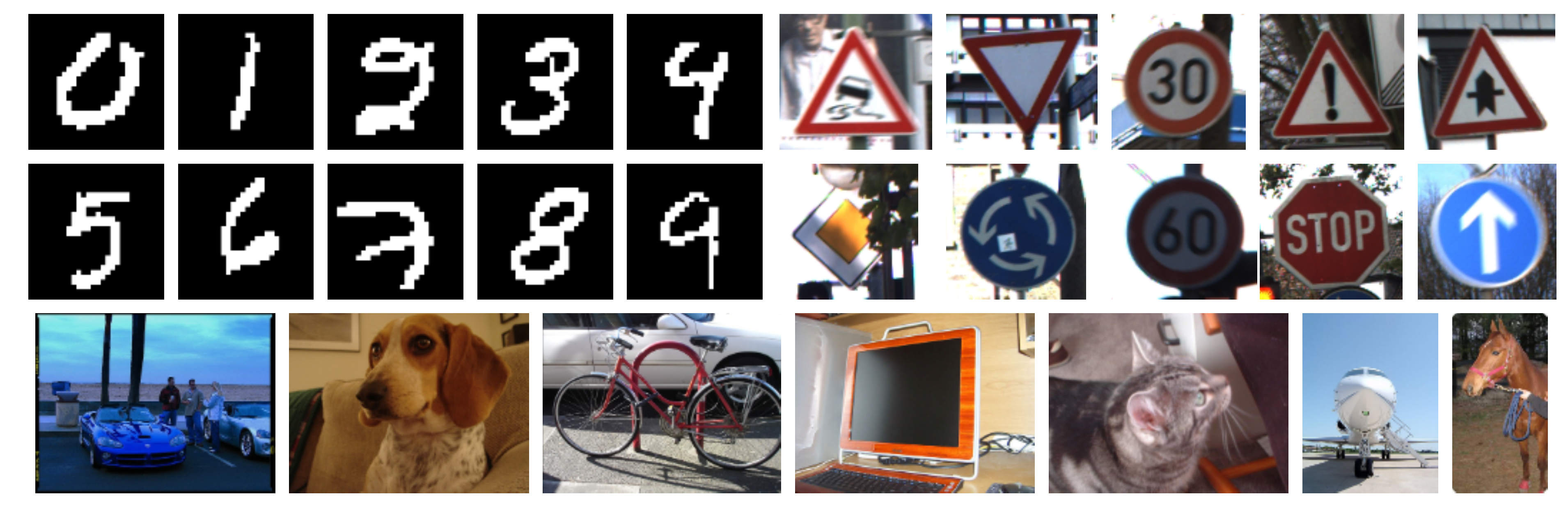}
\caption{
\textbf{Dataset image examples --}
{(top-left)} MNIST. 
{(top-right)} GTSRB. 
{(bottom)} Pascal VOC 2012.
}
\label{fig:data}
\end{figure*}

\section{Sample collapse prevention}
\label{sec:pushaway}
While \eq{linearized_stable} is straightforward to compute, we need to take special care that the random samples do not collapse into a single pixel.
This can happen when the transformation is zooming into a specific region; see \fig{pushaway}~(middle).
If this happens, the data matrix $\bX_i$ generated from coordinate differences of transformed auxiliary samples will be composed of very small numbers, and the data matrix $\bY_i$, generated from difference in intensities, may also become zero.
This leads to exploding gradients.
To avoid them, we perturb the auxiliary samples \textit{after} the transformation; see \fig{pushaway}~(right).
Denoting the modified coordinates for the $k$-th auxiliary sample for pixel $i$ with $\tu_i^k$ and $\tv_i^k$, for $k \in \left\{1,2,...,K\right\}$, we then apply
\begin{align}
    \tu_i^k \leftarrow \tu_i^k + \calN\left(0, \delta_u\right)
    ,
    \\
    \tv_i^k \leftarrow \tv_i^k + \calN\left(0, \delta_v\right)
    ,
\end{align}
where $\delta_u$ and $\delta_v$ correspond to the single pixel width and height in the image that we sample from -- the magnitude of the transform in either dimension.
\section{Results}
To demonstrate the effectiveness of our method, we first present quantitative results on the performance of STN and ICSTN with different sampling methods. 
We show that the enhanced gradients from linearization-based sampling lead to significant improvements in the performance of deep networks, most prominently when downsampling is involved.
We then visually inspect the gradients produced by bilinear sampling and our linearized multi-sampling approach,
to demonstrate that the gradients produced by our method are more robust and lead to better convergence.
We further show that this can be confirmed quantitatively by a simple image alignment task, performed with different sampling methods.
Finally, we study the effect of the number of auxiliary samples and the impact of the magnitude of the sampling noise.

\paragraph{Datasets.}
Our experiments feature three datasets:
\MNIST~\cite{Lecun98}, \TRAFFIC~\cite{Stallkamp11}, and \PASCAL 2012~\cite{Everingham12}; see \fig{data}.
With \MNIST, we show that our approach outperforms bilinear interpolation in one of the classical classification datasets.
With \TRAFFIC, we demonstrate that
our method also leads to better performance in the more challenging case of classifying traffic sign images.
Finally, we use Pascal VOC 2012 to showcase that the advantages of our method are even more prominent on natural images with richer textures.

\paragraph{Baselines.}
In addition to standard bilinear sampling, we also evaluate 
a multi-scale 
baseline where the image is sampled at multiple scales and 
then aggregated.
This is similar to how multi-scale sampling with mipmaps is performed.
To create a multi-scale representation, we create three scale levels, using Gaussian kernels with standard deviations $\{1, 5, 10\}$.
We apply these sampling methods to STN~\cite{Jaderberg15} and ICSTN~\cite{Lin16b} for classification.

\subsection{Implementation details}
We implement our method with PyTorch~\cite{PyTorch}, and use its default bilinear sampling to fetch the intensity corresponding to each of our random samples.
Note that this sampling process is \textit{not} differentiated through, and we use these intensities only to compute the linearization. To ensure this, our implementation explicitly stops the gradient from back-propagating for all variables except for $\transform_{\btheta}\left(\bx\right)$.

To prevent manifestation from out-of-bound samples, we simply mask the sample points that fall outside the original image. 
Specifically, for each pixel $i$, if $\transform_{\btheta}\left(\bx_i^k\right)$ is out of bounds, we exclude it from \eq{data_I} and \eq{data_X}. 
This can be easily achieved by multiplying the corresponding entries with zero.
When all pixels are out of bounds, $\bA_i$ becomes a zero matrix, thus providing a zero gradient.

Throughout our experiments, we use eight auxiliary samples per pixel ($K=8$), a choice we will substantiate in \secref{ablation}.

\newcolumntype{C}{>{\centering\arraybackslash}X}
\begin{table}[t]
\caption{
Test error of models trained with various sampling methods, using STN and ICSTN. 
Best results are marked in bold. 
Our method gives best performance even when there is no downsampling, and the gap widens at higher downsampling rates.
}
\vspace{-1em}
\begin{center}
\small
\begin{tabularx}{\linewidth}{l C C C C}
    \toprule
     Downsampling rate & 1x & 2x & 4x & 8x \\
     \# classif. network param. & 966k & 246k & 61k & 19k \\
     \midrule
     Baseline w/o STN   & 12.37 & 12.88 & 20.85 & 45.88 \\
     \midrule
     STN + Bilinear & 6.29 & 6.50 & 7.95 & 15.31 \\
     STN + Multi-scale & 6.83 & 6.70 & 8.30 & 15.00 \\
     STN + Ours & {\bf 6.08} & {\bf 6.48} & {\bf 7.13} & {\bf 10.89} \\
     \midrule
     ICSTN + Bilinear & 5.68 & 5.00 & 6.52 & 9.80 \\
     ICSTN + Multi-scale & 5.40 & 5.95 & 6.06 & 10.19 \\
     ICSTN + Ours & {\bf 4.85} & {\bf 4.68} & {\bf4.86} & {\bf 6.10} \\
     \bottomrule
\end{tabularx}
\vspace{-1em}
\end{center}
\label{tbl:gtsrb}
\end{table}

\subsection{Classification}

To demonstrate that sampling plays a critical role in networks that contain image transformations, we train a network for image classification with STN/ICSTN prepended to it on the \TRAFFIC dataset.
We emulate a standard setup used to leverage attention mechanisms 
by using STN/ICSTN to produce a transformation of a smaller resolution than the input image, which is then given to the classifier -- the task of STN/ICSTN is thus to focus on a single region that helps to classify traffic signs. We evaluate the accuracy of the classifier under different downsampling rates.
We detail this experiment and the results below.

\paragraph{Network architecture and training setup.}

To train the network we randomly split the training set of \TRAFFIC into two sets, 35309 images for training and 3900 for validation.
For testing we use the provided test set that holds 12630 images.
We crop and resize all images to 50$\times$50.

For the STN/ICSTN module, denoting convolution layers with $c$ channels as $C(c)$, ReLU activations as $R$, and max-pooling layers as $P$, our network is: $C(4)RC(8)RPC(16)RPC(32)RPC(1024)$.
All convolutional layers use 7$\times$7 kernels.
We apply max-pooling over each channel of the last feature map to produce one feature vector of size 1024.
We then apply a fully connected layer with 48 neurons and ReLU activations, followed by another fully connected layer that maps to transformation parameters --  translation, scale, and rotation in our experiments.

For the classification network, we use a simple network with a single hidden layer with 128 neurons and ReLU activations.
We choose a simple architecture on purpose in order to prevent the network relying 
on the increased capacity of a large classification network and learn spatial invariance and effectively ignoring the STN.

\begin{figure}
\centering
\includegraphics[width=\linewidth, trim= 65 0 0 0, clip]{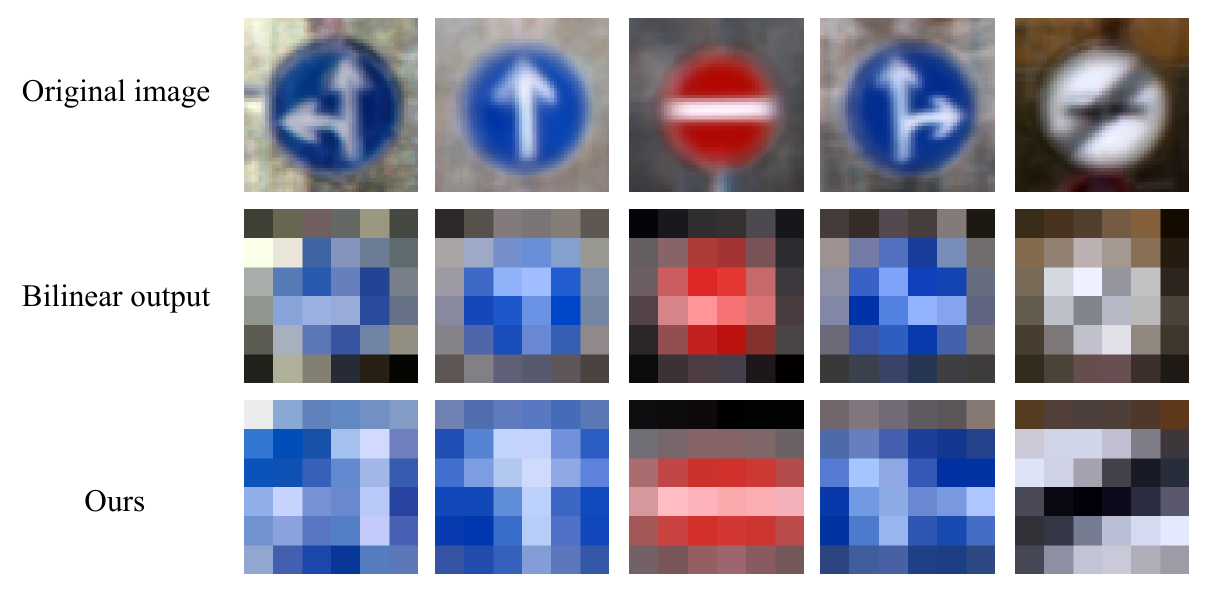}
\caption{
\textbf{Example outputs of trained STN modules -- }
Outputs of STNs trained on the 8x downsampled \TRAFFIC  dataset (top) with bilinear sampling (middle) and our approach (bottom).
The STN trained with our linearized sampling scheme learns to zoom-in for better classification accuracy.
}
\label{fig:stn_qualitative}
\end{figure}

We use ADAM~\cite{Kingma14} as the optimizer, and choose 10$^{-5}$ and 10$^{-3}$ as the learning rates for the STN/ICSTN modules and the classifier respectively.
We train the models from scratch with a batch size of 64, and set the maximum number of iterations to 300k. We use early stopping if the model shows no improvement on the validation split within the last 80k iterations.

\paragraph{Results.}

In \tbl{gtsrb} we show the test accuracy of our approach compared to bilinear sampling and multi-scale sampling, with both STN and ICSTN.
Unsurprisingly, the performance of both degrades as downsampling becomes more severe, but significantly less so with our method.
Notably, the network trained with our sampling method performs the best for \emph{all} downsampling rates -- even without any downsampling.
Moreover, with ICSTN and our method, there is no noticeable performance degradation even with 4x downsampling, allowing  the classification network to be 11 times smaller compared to
when no downsampling is used.

Finally, we show the average 
transformed image for a subset of the classes in the test set in \fig{stn_qualitative}.
Note how the outputs of our network are zoomed-in with respect to the original input image, and more so than the results with bilinear sampling.
This shows that with our linearized sampling strategy the network learns to focus on important regions more effectively.

\def \fitquantw {0.30}
\begin{figure*}
\centering
\subfigure[Downsampling 1x]{
    \includegraphics[width=\fitquantw\linewidth]{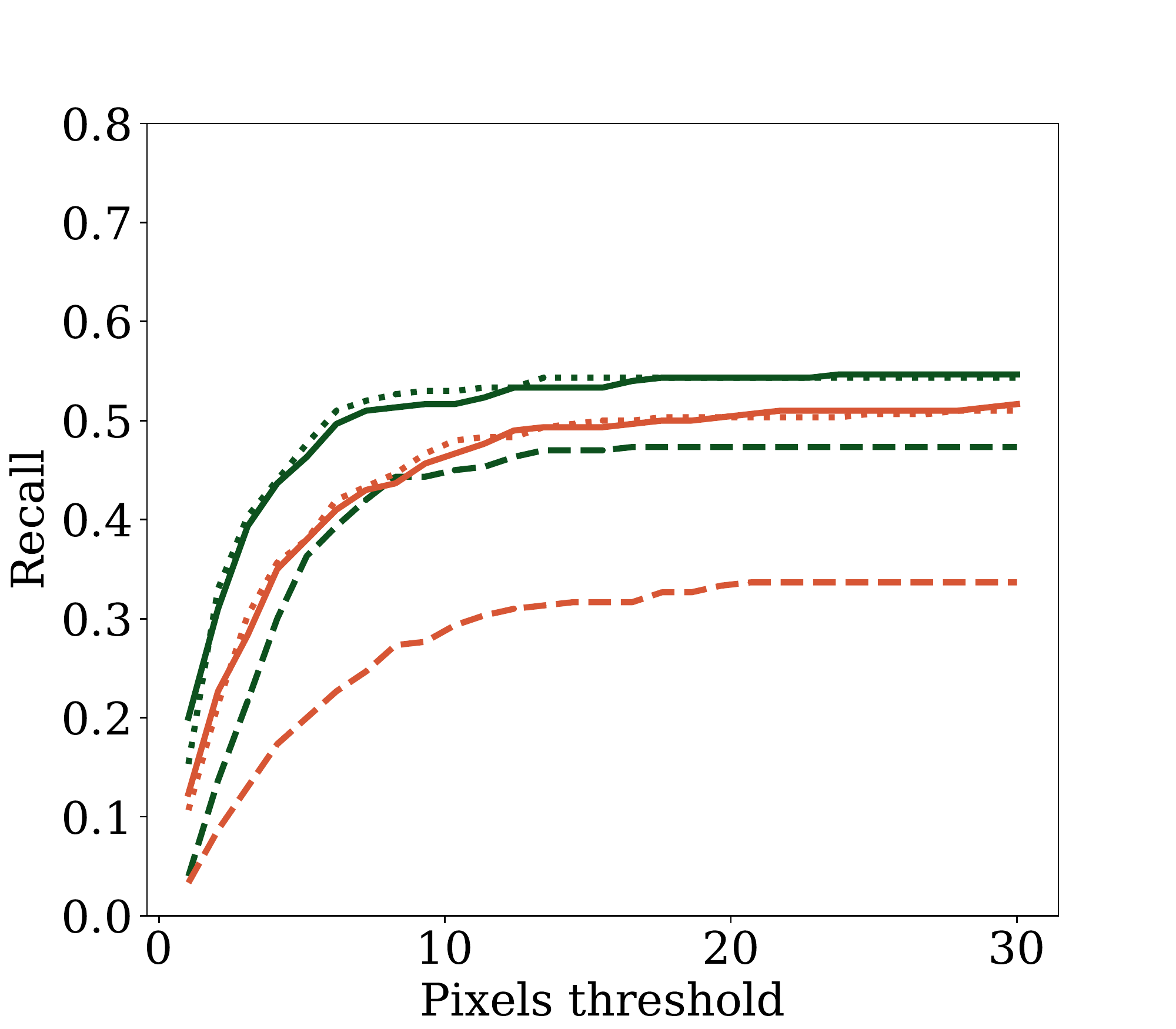}
}
\subfigure[Downsampling 4x]{
    \includegraphics[width=\fitquantw\linewidth]{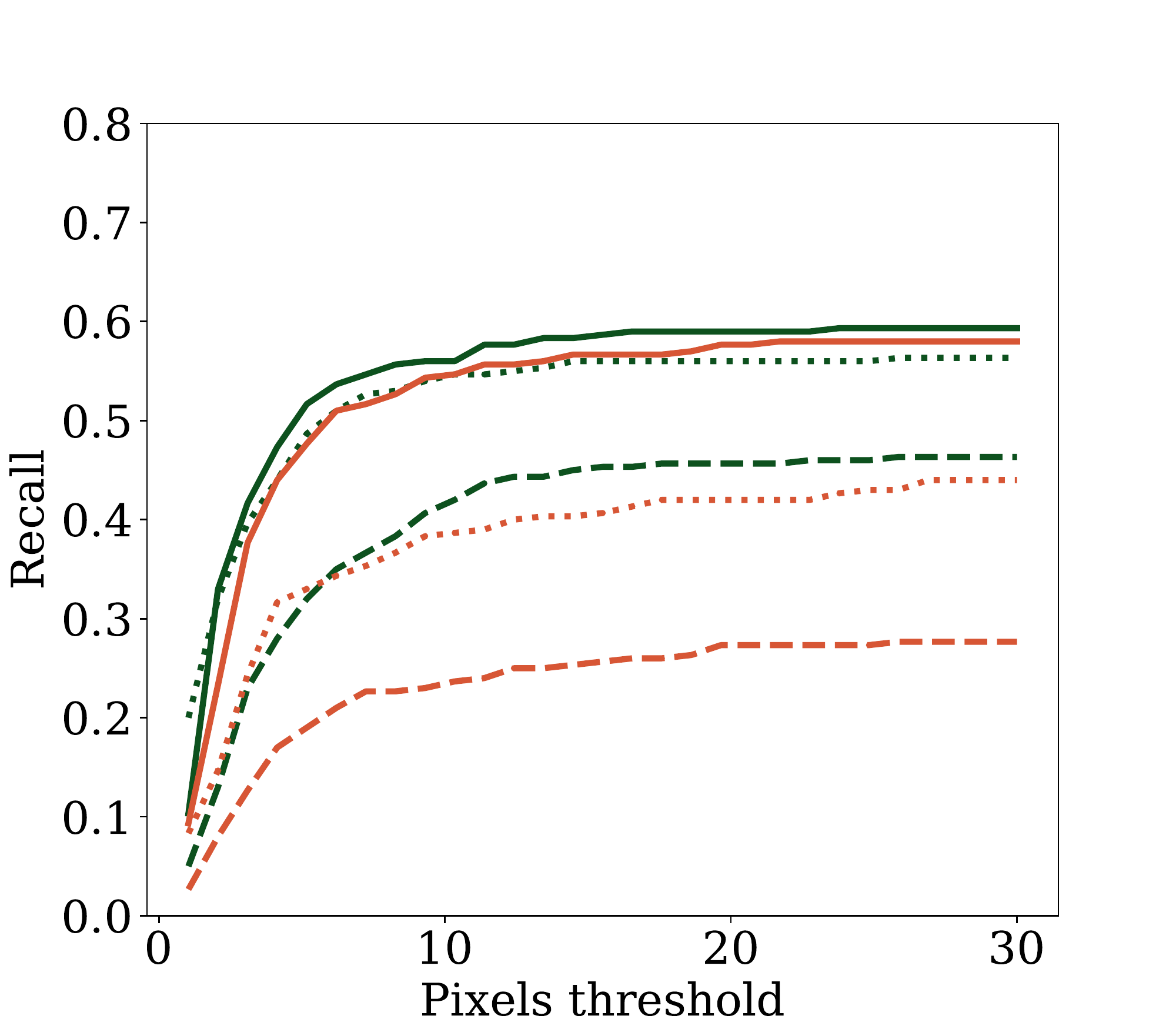}
}
\subfigure[Downsampling 8x]{
    \includegraphics[width=\fitquantw\linewidth]{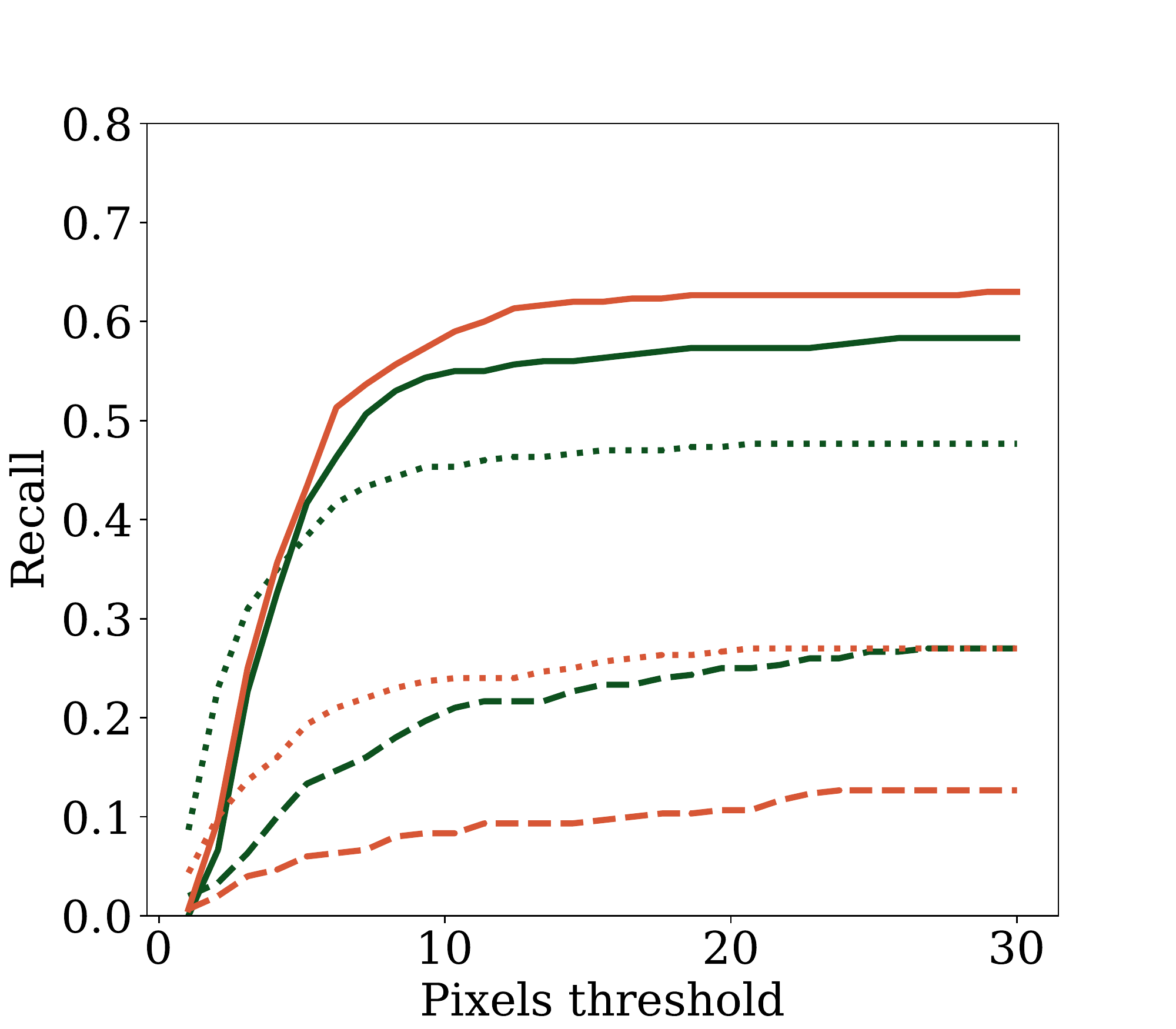}
}
\subfigure{
    \includegraphics[width=0.5\linewidth]{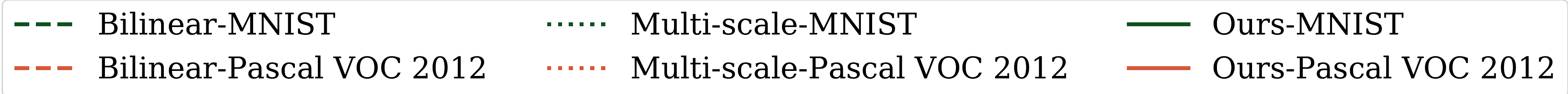}
}
\vspace{-1em}
\caption{
\textbf{Image alignment experiments --} 
We threshold based on the average re-projection error of the four corners of the sampled region. 
Our method (solid line) performs well even under high downsampling rates,
whereas the performance with bilinear sampling (dashed line) and multiscale sampling (dotted line) drops drastically. 
Note also that the gap is even larger for natural images (orange).
}
\label{fig:fit_quantitative}
\end{figure*}
\def \gradfig {0.45}
\begin{figure}
\centering
\includegraphics[width=0.92\linewidth]{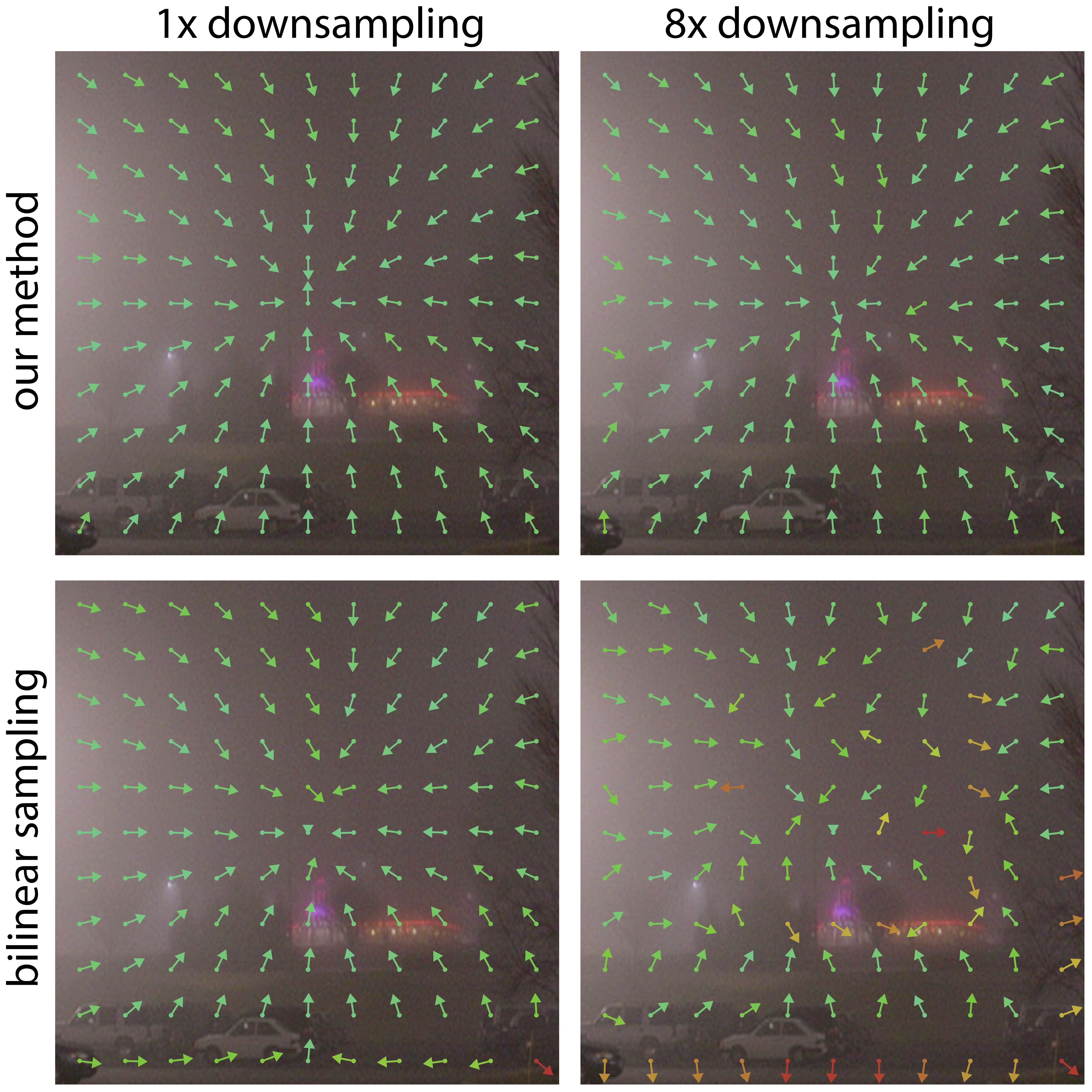}
\caption{
Gradients produced by our method vs bilinear sampling, evaluated across two downsampling factors.
The (negated) image alignment gradients are visualized with arrows, 
colored from green to red according to whether they match the ground truth (all arrows should point towards the center).
Our method shows a wider basin of convergence than bilinear sampling.
The difference becomes more prominent as the downsampling rate increases.
Even without downsampling, we provide better gradients.
}
\label{fig:gradients}
\vspace{-1em}
\end{figure}

\subsection{Gradient analysis}

To understand where such a drastic difference comes from, we show how the gradients produced by our approach differ from those of bilinear sampling.
We compare how the gradients flow when we artificially crop the central region of an image, move it to a different location, and ask the STN to move it back to the original position.
We use natural images from the \PASCAL dataset for this purpose.
Specifically, we crop the central two-thirds of the image, downsample them at different rates with bilinear interpolation, and use this image as our guidance for the STN with an $\ell_2$ loss.

In \fig{gradients} we visualize the negative gradient flows, \ie the direction of steepest descent, extracted on a coarse grid on top of the image for which the gradients were computed.

Since the target area is the central region, the gradients should all point towards the center.
As shown, our method provides a wider basin of convergence, having the gradients consistently point towards the center, \ie, the ground truth location, from points further away than with bilinear sampling.
Note that while this has a drastic effect for the case of downsampling by 8x, it holds true even when there is no downsampling (\fig{gradients}, left).
This is
particularly distinctive around the bottom of the image, which has richer structures than the rest, which is blurry and thus rather smooth.
This causes bilinear sampling to provide poor gradients, whereas our method is not susceptible to this.

\def \ablsigmaw {0.23}
\def \ablsigmah {0.18}
\begin{figure*}
\centering
\subfigure[Recall vs threshold]{
    \includegraphics[width=\ablsigmaw\linewidth, height=\ablsigmah\linewidth, trim=0 0 0 0, clip]{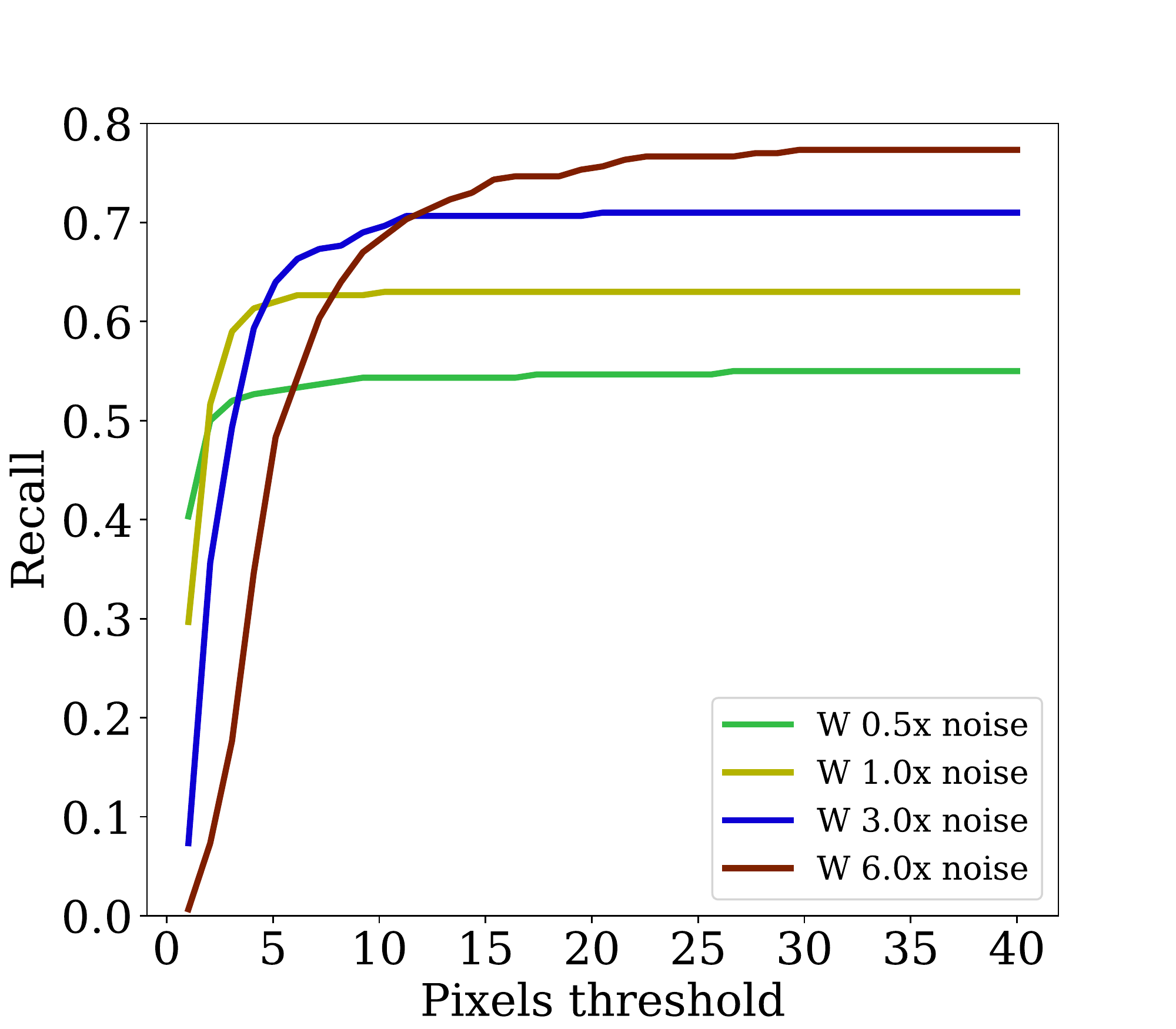}
}
\subfigure[$\bsigma=1$px]{
    \includegraphics[width=\ablsigmaw\linewidth, height=\ablsigmah\linewidth, trim = 0 80 0 20, clip]{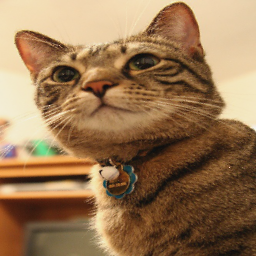}
}
\subfigure[$\bsigma=3$px]{
    \includegraphics[width=\ablsigmaw\linewidth, height=\ablsigmah\linewidth, trim = 0 80 0 20, clip]{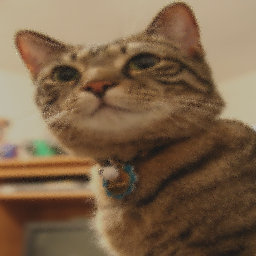}
}
\subfigure[$\bsigma=6$px]{
    \includegraphics[width=\ablsigmaw\linewidth, height=\ablsigmah\linewidth, trim = 0 80 0 20, clip]{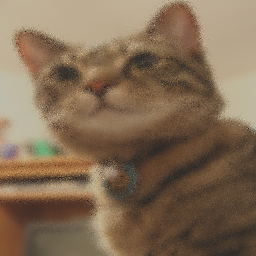}
}
\caption{
\textbf{Ablation test for $\bsigma$ --}
Effect of the magnitude of the noise used to generate the auxiliary samples. ({\bf a}) Quantitative results in terms of recall vs threshold. ({\bf b-d}) The spatial context increases along with the noise (left to right), as the auxiliary samples spread out, as indicated by the blurring.
In ({\bf a}), more noise results in more spatial context thus a larger basin of convergence, but at the cost of final accuracy.
Also, as shown in ({\bf b-d}), the image becomes more blurry as compromise.
}
\label{fig:ablation_sigma}
\end{figure*}

\def \ablnumw {0.47}
\begin{figure}
\centering
\subfigure[]{
    \includegraphics[width=\ablnumw\linewidth]{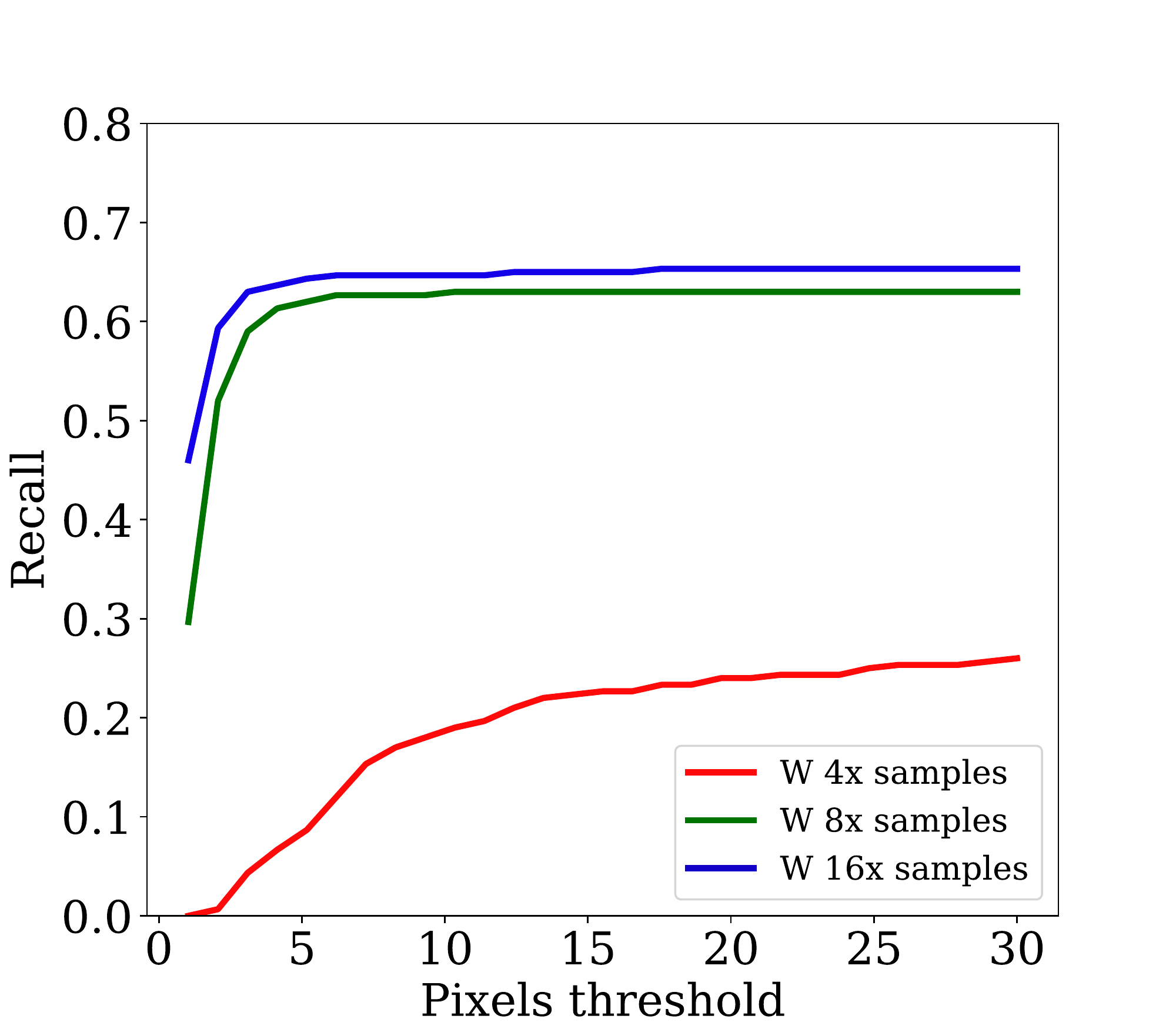}
}
\subfigure[]{
    \includegraphics[width=\ablnumw\linewidth]{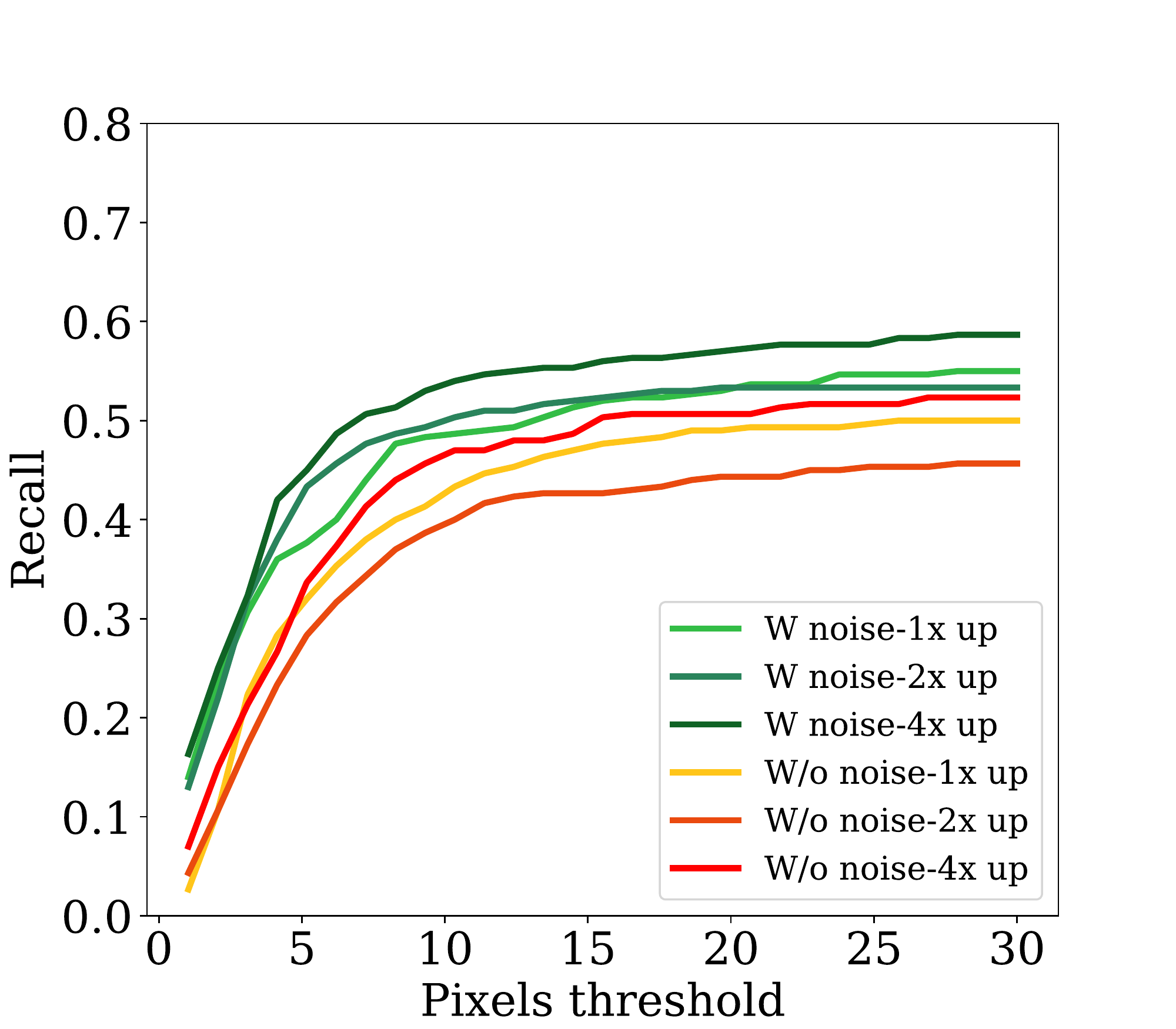}
}
\vspace{-1em}
\caption{
\textbf{Ablation test on the number of samples --}
Effect of ({\bf a}) number of samples on image alignment under 4x downsampling, and ({\bf b}) sample collapse prevention under 1x, 2x, and 4x upsampling.
Both are in terms of recall vs threshold.
We again use the average reprojection error of four corners of the sampled region.
In ({\bf a}), increasing the number of samples provides better results in terms of alignment.
In ({\bf b}), we demonstrate that sample collapse prevention is essential when upsampling.}
\label{fig:ablation_number_and_second}
\end{figure}

\subsection{Optimizing for image alignment}

Next, we demonstrate that this improvement in the gradients produced by our approach leads to better outcomes when \textit{aligning} images, as illustrated earlier in \fig{teaser}.
In order to isolate the effect of the sampler, we exclude the classification network and directly optimize the parameters of a STN for one image at a time, with a single synthetic perturbation. We then evaluate the quality of the alignment after convergence.
We repeat this experiment 80 times, on randomly selected images from the different datasets, and with random transformations.

For the perturbations, we apply random in-plane rotations, scale changes (represented in log-space, \ie scale of 0.5 is represented as $-1$) in the horizontal and vertical directions (independently), and translations. 
We sample Gaussian noise with standard deviation of 1 and apply it as follows.
With the image coordinates normalized to $[-1,1]$, we use a standard deviation of ${\pi}/{4}$ for rotations, $\sqrt{2}$ for scale changes 
in horizontal and vertical directions separately, and $0.2$ for translations.

We summarize our image alignment experiments in \fig{fit_quantitative}.
As shown, our results are significantly better than those of bilinear sampling, even when there is no downsampling.
We also outperform the multi-scale if there is any downsampling at all.
More importantly, the gap between methods is specially wide on natural images, where the image characteristics are much more complex than for the other two datasets.
Finally, the effects of downsampling become quite severe at 8x, effectively breaking the bilinear sampler and the multi-scale approach in all datasets, and for PASCAL VOC in particular.

\subsection{Ablation tests}
\label{sec:ablation}

\paragraph{Auxiliary sample noise.}

We also examine the effect of the magnitude of the noise we use to place the auxiliary samples in \eq{sample}.
\fig{ablation_sigma} (b-c) shows the sampling result under various $\bsigma$.
Since the Gaussian noise emulates a point spread function, with a larger $\bsigma$ we obtain a blurrier image via \eq{interp}.
While the outcome is blurry, this effectively allows
our method to consider a wider support region in the computation of the gradients, which will therefore be smoother.
As shown in \fig{ablation_sigma}~(a), the wider spatial extent covered by the random samples translates into better image alignment results in the \PASCAL dataset. 

\paragraph{Number of auxiliary samples.}

In \fig{ablation_number_and_second}~(a)
we evaluate the effect of number of random samples taken for each pixel with the \PASCAL dataset under 4x downsampling.
As expected, increasing the number of auxiliary samples leads to better alignment results.
When only four auxiliary samples are used, the linearization is quite ill-conditioned and thus accuracy of the method drops.

\paragraph{Sample collapse avoidance.}

In 
\fig{ablation_number_and_second}~(b)
we demonstrate the importance of the sample collapse prevention scheme described in \refsec{pushaway}.
When our technique for sample collapse prevention is removed, 
the performance in image alignment drops due to the numerical instability caused by all auxiliary samples falling in
the same neighbourhood.
By contrast, our approach prevents this problem.

\section{Conclusions and future work}

We have proposed a linearization method based on multi-sampling to improve the poor gradients provided by bilinear sampling -- a de-facto standard for differentiable image manipulation, typically within Spatial Transformers -- for example, to implement hard attention.
We have demonstrated empirically that our method can provide improved gradients, leading to enhanced performance on downstream tasks.
Our approach simply swaps the sampler used by networks that rely on Spatial Transformers, and is thus compatible with any improvements that build on top of them. 

Note that it is possible to replace the sampler with bilinear interpolation at inference time to preserve efficiency.
As of now, this would result in a minor change in the sampling outcome and require fine tuning of the classifier.
As immediate future work, we are currently investigating a bias-free formulation for \eq{data_X}, which would allow in-place replacement without further fine tuning.

\ificcvfinal
\section*{Acknowledgements}

This work was partially supported by the Natural Sciences and Engineering Research Council of Canada Discovery Grant ``Deep Visual Geometry Machines'' (RGPIN-2018-03788, DGECR-2018-00426), Google, and by systems supplied by Compute Canada. 

\fi

{\small
\bibliographystyle{ieee_fullname}
\bibliography{full}
}
\newpage
\cleardoublepage
\onecolumn
\appendix

\section{Supplementary appendix}

\subsection{Computation time}
The amount of overhead depends on the number of auxiliary samples, and the size of the input images. The increase is mainly in the backward pass. With default parameters, our method runs on average at a speed of 67\% compared to bilinear sampling for the forward pass, and 40\% for the backward pass.

\subsection{Results on MNIST}
As MNIST dataset is textureless,
both sampling methods perform similarly -- we used the implementation of~\cite{Lin16b} and achieved 1.4\% error rate for both bilinear sampling and ours.
While the two perform similar, as shown on the average aligned image below, ours zooms in more on digits.
\def \qualw {0.99}
\begin{figure}[h]
\centering
\includegraphics[width=\qualw\linewidth, trim = 0 0 0 0, clip]{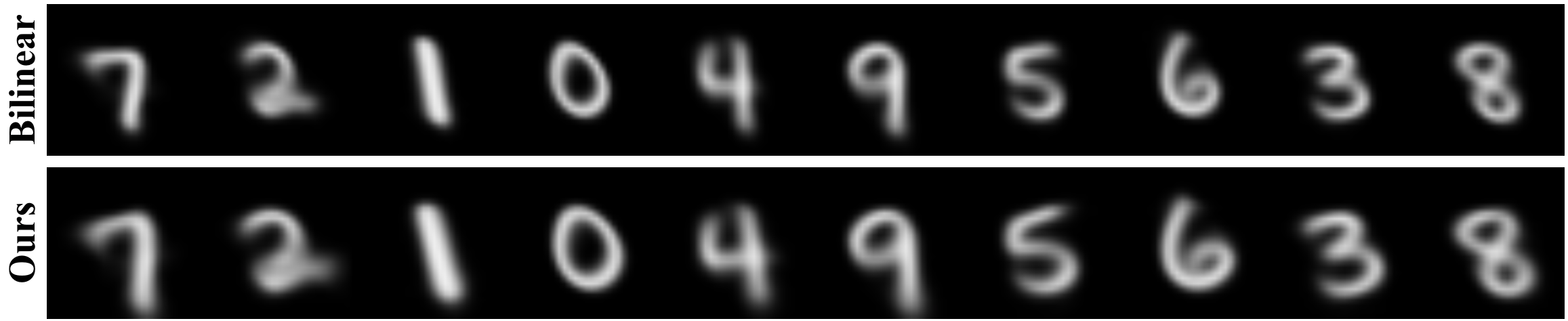}
\caption{
Average 
aligned image from the STN using (top) bilinear sampling and (bottom) our method.
\vspace{-1em}
}
\label{fig:qual}
\end{figure}

\end{document}